\documentclass{article}
\usepackage{spconf,amsmath,graphicx,hyperref,float,booktabs,multirow}
\usepackage[table]{xcolor}

\definecolor{tablesection}{gray}{0.93}
\definecolor{tableaverage}{gray}{0.965}

\title{SynLat: Syntax-Aligned Text--Latent Compression for Chain-of-Thought Reasoning}
\name{\shortstack{Yifeng Zhao$^{1}$, Hongjun Yu$^{1}$, Shibo Wang$^{1}$,
Yunjiao Zhou$^{1}$, Zixiao Zhu$^{1}$,\\
Zepeng Ning$^{2}$, Kezhi Mao$^{1}$, Junlang Qian$^{1,*}$}\thanks{$^{*}$Corresponding author: \href{mailto:JUNLANG001@e.ntu.edu.sg}{JUNLANG001@e.ntu.edu.sg}}}
\address{$^{1}$Nanyang Technological University \qquad
$^{2}$Nankai University}
\begin{document}
\ninept
\renewcommand{\baselinestretch}{0.941}\normalsize 
\setlength{\parskip}{0pt}
\setlength{\textfloatsep}{10pt}
\maketitle
\begin{abstract}
Long chain-of-thought (CoT) traces impose substantial output-token costs. Under
constrained budgets, compression must preserve answer-critical information,
making boundary placement central. Token-level and fixed-length boundaries can
fragment coherent spans such as phrases, formulas, and local derivations,
whereas step-level boundaries can bind content requiring different compression
actions. We introduce SynLat, a text--latent CoT framework that aligns
compression boundaries with syntactic structure through non-overlapping
Syntax-Aligned Units (SAUs). An answer-conditioned Teacher constructs
progressive KEEP/LATENT targets for a single compression-conditioned Student,
which generates mixed reasoning from only the question and requested
compression level at inference. Across two Qwen3 Student scales, Standard- and
Long-CoT groups, and three compression levels, SynLat matches or exceeds the
strongest evaluated baseline in all 12 task-group aggregates and strictly leads in 11
under the reported achieved-CR selection protocol. Overall gains reach 3.6/2.6
points at MEDIUM and 7.0/5.5 points at HIGH for Qwen3-8B/14B, with larger
advantages under stronger compression, particularly on Long-CoT groups.
\end{abstract}
\section{Introduction}
\label{sec:intro}

Chain-of-thought (CoT) prompting makes intermediate reasoning explicit
\cite{wei2022chain}, but long traces increase output-token cost. Explicit
compression methods include token pruning \cite{xia2025}, chunk compression
\cite{wang2025}, step-entropy filtering \cite{li2026b}, and formula
preservation \cite{tang2026}. At aggressive ratios, however, deleting or
condensing text can reduce logical fidelity and fragment symbolic expressions
\cite{tang2026}.

\begin{figure}[t]
  \centering
  \includegraphics[width=\columnwidth,trim=4pt 3pt 4pt 3pt,clip]{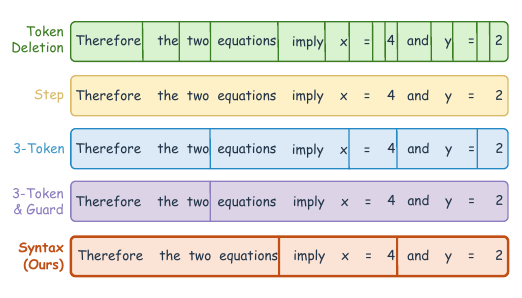}
  \vspace{-4pt}
  \caption{Compression-decision boundaries for the same reasoning sentence.
  Token and fixed-window units can fragment local relations, whereas step-level
  units can bind distinct relations; Syntax-Aligned Units preserve coherent
  relations while keeping them separately assignable to text or latent form.}
  \label{fig:compression-boundaries}
\end{figure}

Latent compression instead moves reasoning into non-text carriers. COCONUT
propagates hidden states as continuous reasoning states \cite{hao2025coconut},
whereas CODI distills explicit CoT into continuous space \cite{shen2025codi}.
With the trace fully replaced, latent states must carry complete, precise
derivations without textual anchors. Moreover, SIM-CoT reports that scaling
implicit reasoning can destabilize training as latent representations become
homogeneous and lose semantic diversity \cite{wei2025simcot}. Hybrid methods
therefore retain selected text: Token Assorted mixes it with discrete latent
symbols \cite{su2025token}, while SPOT replaces selected spans with pause states
\cite{chu2026spot}. Compression then becomes a representation-fidelity
allocation problem: which content remains textual and which becomes latent.

However, this allocation requires suitable decision units, and existing methods
expose a boundary mismatch. As Figure~\ref{fig:compression-boundaries}
illustrates, token and fixed-window
units can fragment local relations, whereas step-level units can bind distinct
relations into one compression decision. Neither provides trace-specific units
that keep local relations intact while leaving them separately assignable to
text or latent form.

Constituency syntax supplies nested contiguous spans for such allocation units
\cite{kitaev2018constituency}. SynLat converts them into ordered,
non-overlapping Syntax-Aligned Units (SAUs). Offline, an answer-conditioned
Teacher assigns progressive \texttt{KEEP}/\texttt{LATENT} targets, and latent
units map to a shared discrete codebook. Plan-then-Generate training transfers
this supervision to a compression-conditioned Student. At inference, the
Student receives only the question and requested compression level, generating
a mixed CoT of retained text and latent symbols before the final answer.

Experiments show that, under the reported achieved compression rate (CR)
selection protocol, SynLat improves the accuracy--compression trade-off, with
larger gains under stronger compression, particularly on Long-CoT groups.
Controlled studies further support SAU boundaries, answer-conditioned planning,
and Student-side transfer.

Our contributions are threefold: (i) we formulate budgeted CoT compression as a
representation-fidelity allocation problem, retaining answer-critical reasoning
as text while compressing recoverable content into latent representations; (ii)
we introduce SynLat, which aligns text--latent allocation boundaries with
trace-specific syntax and transfers answer-conditioned targets to a single
Student; and (iii) experiments establish its accuracy--compression gains across
the evaluated model scales, task groups, and compression levels under the
reported achieved-CR selection protocol.

\begin{figure*}[t]
  \centering
  \includegraphics[width=\textwidth,trim=4pt 3pt 4pt 3pt,clip]{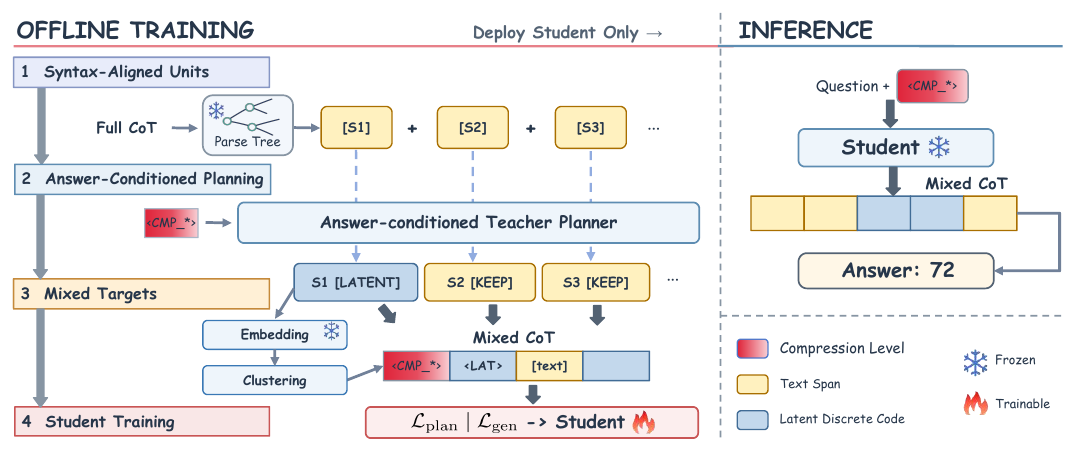}
  \vspace{-4pt}
  \caption{Overview of SynLat. Offline target construction converts a
  reference trace into mixed-CoT targets through SAU construction, KEEP/LATENT
  allocation, and latent-code assignment; at deployment, the Student uses only
  the question and compression condition.}
  \label{fig:framework-overview}
\end{figure*}

\section{RELATION TO PRIOR WORK}
\label{sec:prior}

\noindent\textbf{Explicit CoT and visible compression.}
Chain-of-thought prompting exposes intermediate reasoning
\cite{wei2022chain}, whose growing traces motivate compression within natural
language. TokenSkip prunes tokens \cite{xia2025}; R1-Compress condenses chunks
\cite{wang2025}; step-entropy filtering selects steps by entropy
\cite{li2026b}; and Extra-CoT preserves formulas at aggressive ratios
\cite{tang2026}. These methods remain within visible text and
delete or rewrite predefined units, motivating representations beyond text.

\noindent\textbf{Latent and hybrid text--latent reasoning.}
Latent methods move reasoning beyond ordinary text. COCONUT propagates
continuous hidden states \cite{hao2025coconut}, while CODI distills explicit
CoT into continuous space \cite{shen2025codi}. SIM-CoT reports that scaling
implicit reasoning can destabilize training and homogenize latent states
\cite{wei2025simcot}. DLR learns a discrete vocabulary from rendered CoT features
\cite{chang2026dlr}. Hybrid methods therefore retain textual anchors: Token
Assorted mixes text with discrete symbols \cite{su2025token}, SPOT replaces
spans with pause states \cite{chu2026spot}, Selective Latent Thinking gates
latent spans by confidence \cite{xie2026selective}, and A*-Thought-V2
geometrically selects step-level continuous states \cite{xu2026astar}. Prior
summarization work uses constituency parses to define deletable spans
\cite{xu2019constituency}. SynLat differs by flattening trace-specific syntax
into ordered, non-overlapping SAUs and learning compression-conditioned
text--latent allocation over a shared discrete codebook.

\section{METHOD}
\label{sec:method}

\subsection{Preliminaries}

Given a question $q$, standard chain-of-thought (CoT) reasoning generates a
textual trace $R=(w_1,\ldots,w_n)$ before an answer $y$, i.e.,
$q\rightarrow R\rightarrow y$, where each $w_i$ is a text token. In our
discrete setting, a consecutive segment $w_{a_j:b_j}$ maps to a latent token
$z_j=\phi(w_{a_j:b_j})$, shortening the trace while discarding its
natural-language form. We study mixed latent CoT, written as
$\tilde R=(u_1,\ldots,u_k)$, where each $u_i$ is a retained text span or latent
token, e.g.,
\[
\tilde R=(w_{1:a},z_1,w_{b:d},z_2,\ldots).
\]

\begin{table*}[!t]
\centering
\caption{Main results. Each cell reports answer accuracy and achieved
compression rate (Acc./CR); Avg. and Overall are macro averages. Bold accuracy
is best within each compression setting, and bold method names indicate
SynLat. Raw Full CoT has no normalized CR; $\dagger$ marks the nearest point
outside the strict budget-matching window.}
\label{tab:main-results}
\setlength{\tabcolsep}{2.6pt}
\renewcommand{\arraystretch}{1.10}
\setlength{\aboverulesep}{0.35ex}
\setlength{\belowrulesep}{0.5ex}
\resizebox{\textwidth}{!}{%
\begin{tabular}{llcc>{\columncolor{tableaverage}}cccc>{\columncolor{tableaverage}}c>{\columncolor{tableaverage}}c}
\toprule
& & \multicolumn{3}{c}{\textbf{Standard-CoT}} & \multicolumn{4}{c}{\textbf{Long-CoT}} & \cellcolor{white} \\
\cmidrule(lr){3-5}\cmidrule(lr){6-9}
\textbf{Setting} & \textbf{Method} & \shortstack{\textbf{MATH-500}\\Acc./CR} &
\shortstack{\textbf{StrategyQA}\\Acc./CR} & \shortstack{\textbf{Avg.}\\Acc./CR} &
\shortstack{\textbf{OlympiadBench}\\Acc./CR} & \shortstack{\textbf{WorldTree V2}\\Acc./CR} &
\shortstack{\textbf{ProofWriter}\\Acc./CR} & \shortstack{\textbf{Avg.}\\Acc./CR} &
\shortstack{\textbf{Overall}\\Acc./CR} \\
\midrule
\rowcolor{tablesection}\multicolumn{10}{l}{\textbf{(a) Qwen3-8B Student}} \\
\midrule
\multirow{3}{*}{Reference} & No CoT & 64.1/100.0 & 64.8/100.0 & 64.5/100.0 & 12.6/100.0 & 68.9/100.0 & 75.4/100.0 & 52.3/100.0 & 57.2/100.0 \\
& Raw Full CoT & 87.2/-- & 75.0/-- & 81.1/-- & 49.4/-- & 83.3/-- & 92.7/-- & 75.1/-- & 77.5/-- \\
& \textbf{SynLat (NONE)} & 87.8/0.0 & 75.4/0.0 & 81.6/0.0 & 49.8/0.0 & 83.6/0.0 & 93.1/0.0 & 75.5/0.0 & 77.9/0.0 \\
\midrule
\multirow{4}{*}{LOW} & TokenSkip & 84.9/23.5 & 72.8/21.5 & 78.9/22.5 & 44.5/25.4 & 80.3/22.8 & 89.4/23.0 & 71.4/23.7 & 74.4/23.2 \\
& Token Assorted & \textbf{85.2}/22.7 & 73.1/20.8 & \textbf{79.2}/21.8 & 44.0/24.8 & 80.8/21.9 & 89.9/22.4 & 71.6/23.0 & 74.6/22.5 \\
& SPOT & 84.6/21.8 & 73.4/22.1 & 79.0/22.0 & 44.8/20.2$^\dagger$ & 81.3/23.2 & 90.1/22.8 & 72.1/22.1 & 74.8/22.0 \\
& \textbf{SynLat (LOW)} & 84.8/23.6 & \textbf{73.5}/20.1 & \textbf{79.2}/21.9 & \textbf{46.8}/26.0 & \textbf{81.4}/21.7 & \textbf{91.2}/22.5 & \textbf{73.1}/23.4 & \textbf{75.5}/22.8 \\
\midrule
\multirow{4}{*}{MEDIUM} & TokenSkip & 80.6/45.2 & 68.8/42.8 & 74.7/44.0 & 37.0/47.7 & 74.3/44.1 & 83.0/44.8 & 64.8/45.5 & 68.7/44.9 \\
& Token Assorted & 81.1/43.6 & 69.4/40.2 & 75.3/41.9 & 35.6/44.0$^\dagger$ & 75.7/41.7 & 84.5/42.3 & 65.3/42.7 & 69.3/42.4 \\
& SPOT & 79.6/42.6 & 69.9/44.0 & 74.8/43.3 & 36.2/37.8$^\dagger$ & 76.8/45.0 & 85.4/44.6 & 66.1/42.5 & 69.6/42.8 \\
& \textbf{SynLat (MEDIUM)} & \textbf{83.3}/46.2 & \textbf{71.4}/40.2 & \textbf{77.4}/43.2 & \textbf{43.5}/49.1 & \textbf{79.2}/42.4 & \textbf{88.8}/43.8 & \textbf{70.5}/45.1 & \textbf{73.2}/44.3 \\
\midrule
\multirow{4}{*}{HIGH} & TokenSkip & 73.8/65.8 & 63.2/62.2 & 68.5/64.0 & 25.5/68.9 & 67.0/65.0 & 76.0/66.1 & 56.2/66.7 & 61.1/65.6 \\
& Token Assorted & 74.5/63.1 & 63.8/57.8 & 69.2/60.5 & 21.8/61.5$^\dagger$ & 72.5/60.2 & 77.0/61.0 & 57.1/60.9 & 61.9/60.7 \\
& SPOT & 70.5/61.2$^\dagger$ & 68.0/64.0$^\dagger$ & 69.3/62.6 & 19.5/54.5$^\dagger$ & 73.8/65.8 & 79.7/64.9 & 57.7/61.7 & 62.3/62.1 \\
& \textbf{SynLat (HIGH)} & \textbf{79.1}/66.8 & \textbf{68.7}/58.4 & \textbf{73.9}/62.6 & \textbf{38.7}/71.2 & \textbf{75.0}/62.6 & \textbf{84.8}/64.1 & \textbf{66.2}/66.0 & \textbf{69.3}/64.6 \\
\midrule
\rowcolor{tablesection}\multicolumn{10}{l}{\textbf{(b) Qwen3-14B Student}} \\
\midrule
\multirow{3}{*}{Reference} & No CoT & 67.2/100.0 & 66.1/100.0 & 66.7/100.0 & 15.9/100.0 & 71.8/100.0 & 78.0/100.0 & 55.2/100.0 & 59.8/100.0 \\
& Raw Full CoT & 89.6/-- & 77.4/-- & 83.5/-- & 53.4/-- & 85.0/-- & 94.3/-- & 77.6/-- & 79.9/-- \\
& \textbf{SynLat (NONE)} & 90.2/0.0 & 77.1/0.0 & 83.7/0.0 & 52.7/0.0 & 86.0/0.0 & 95.0/0.0 & 77.9/0.0 & 80.2/0.0 \\
\midrule
\multirow{4}{*}{LOW} & TokenSkip & 87.7/23.0 & 75.0/22.0 & 81.4/22.5 & 48.1/24.7 & 83.0/22.6 & 92.0/23.4 & 74.4/23.6 & 77.2/23.1 \\
& Token Assorted & 87.2/22.5 & 74.9/21.0 & 81.1/21.8 & 47.2/24.0 & 82.6/22.0 & 91.5/22.7 & 73.8/22.9 & 76.7/22.4 \\
& SPOT & 86.9/22.0 & 75.3/22.5 & 81.1/22.3 & 47.8/21.5 & 83.3/23.5 & \textbf{92.2}/23.0 & 74.4/22.7 & 77.1/22.5 \\
& \textbf{SynLat (LOW)} & \textbf{88.7}/24.0 & \textbf{75.9}/20.5 & \textbf{82.3}/22.3 & \textbf{50.2}/25.0 & \textbf{84.2}/22.0 & 91.0/23.0 & \textbf{75.1}/23.3 & \textbf{78.0}/22.9 \\
\midrule
\multirow{4}{*}{MEDIUM} & TokenSkip & 84.9/45.4 & 72.2/44.0 & 78.6/44.7 & 42.9/46.8 & 78.8/44.7 & 88.1/45.1 & 69.9/45.5 & 73.4/45.2 \\
& Token Assorted & 84.0/44.2 & 71.6/41.0 & 77.8/42.6 & 40.1/44.5 & 78.1/42.5 & 87.0/43.1 & 68.4/43.4 & 72.2/43.1 \\
& SPOT & 82.7/43.7 & \textbf{74.0}/44.5 & 78.4/44.1 & 40.8/40.5$^\dagger$ & 79.6/45.2 & 88.7/45.0 & 69.7/43.6 & 73.2/43.8 \\
& \textbf{SynLat (MEDIUM)} & \textbf{86.6}/46.0 & 73.8/40.5 & \textbf{80.2}/43.3 & \textbf{47.1}/47.5 & \textbf{82.0}/43.0 & \textbf{90.6}/44.5 & \textbf{73.2}/45.0 & \textbf{76.0}/44.3 \\
\midrule
\multirow{4}{*}{HIGH} & TokenSkip & 80.0/66.0 & 65.0/64.0$^\dagger$ & 72.5/65.0 & 35.0/67.2 & 73.5/65.4 & 82.0/66.0 & 63.5/66.2 & 67.1/65.7 \\
& Token Assorted & 78.0/64.2 & 63.4/59.0 & 70.7/61.6 & 29.0/62.3$^\dagger$ & 71.0/61.5 & 80.0/62.2 & 60.0/62.0 & 64.3/61.8 \\
& SPOT & 74.2/62.0$^\dagger$ & 67.0/64.8$^\dagger$ & 70.6/63.4 & 24.2/58.0$^\dagger$ & 75.0/66.0 & 85.0/65.5 & 61.4/63.2 & 65.1/63.3 \\
& \textbf{SynLat (HIGH)} & \textbf{83.3}/67.0 & \textbf{71.2}/59.0 & \textbf{77.3}/63.0 & \textbf{43.0}/69.0 & \textbf{78.5}/63.0 & \textbf{87.2}/65.0 & \textbf{69.6}/65.7 & \textbf{72.6}/64.6 \\
\bottomrule
\end{tabular}%
}
\end{table*}

\subsection{Syntactic constituency structure}

\begingroup
\setlength{\intextsep}{6pt}
\begin{figure}[H]
  \centering
  \includegraphics[width=\columnwidth,trim=4pt 5pt 4pt 5pt,clip]{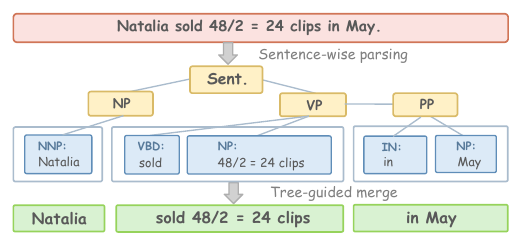}
  \vspace{-4pt}
  \caption{Syntactic constituency parsing and SAU construction. A reasoning
  sentence is parsed into nested contiguous spans, then partitioned into
  non-overlapping Syntax-Aligned Units for text--latent allocation.}
  \label{fig:sau-construction}
\end{figure}
\endgroup

Constituency parsing represents sentences as nested contiguous spans
\cite{kitaev2018constituency}.
Applied sentence-wise, a parser maps $R$ to trees
$T_s=(V_s,E_s)$ with token leaves and internal nodes $v\in V_s$ covering
spans $C_v=w_{a(v):b(v)}$. SynLat uses these boundaries rather than
syntactic labels to define allocation units finer than full reasoning steps.
As Figure~\ref{fig:sau-construction} illustrates, the hierarchical,
overlapping spans are flattened into ordered, non-overlapping
Syntax-Aligned Units (SAUs).

\subsection{SynLat: syntax-aligned text--latent CoT compression}

\noindent\textbf{Overview.}
As Figure~\ref{fig:framework-overview} illustrates, SynLat constructs
mixed-CoT targets offline from a reference trace $R$ and
answer $y^*$ through SAU construction, KEEP/LATENT allocation, and code
assignment. Compression condition $c$ controls how many units are replaced;
the Teacher selects which ones, and two-stage training transfers these targets
to a Student.

\noindent\textbf{Tree-guided syntax-aligned unit construction.}
Starting at each parse root, SynLat retains nodes of at most $L_{\max}$
tokens and otherwise descends to their children. Adjacent siblings sharing a
parent are merged within the same cap; residual content falls back to leaf
spans. Sorting retained spans by their character intervals yields
$S_{1:N}=(S_1,\ldots,S_N)$, an ordered, non-overlapping partition that
fully reconstructs $R$.

\noindent\textbf{Compression-conditioned progressive allocation.}
For each $S_i$, SynLat assigns
$a_i^{(c)}\in\{\mathrm{KEEP},\mathrm{LATENT}\}$: KEEP preserves its text,
whereas LATENT replaces it with a discrete token. Condition $c$ specifies a
target SAU replacement ratio $r(c)$, giving $K_c=\lfloor r(c)N\rfloor$
replacements and $A_c=\{i\mid a_i^{(c)}=\mathrm{LATENT}\}$ with
$|A_c|=K_c$. Unlike achieved token-level CR, this target counts units.
Using $y^*$, the offline Teacher retains answer-critical text and replaces
units it judges recoverable from the remaining textual anchors and context, rather than
judging local importance alone.

The conditions share a progressive replaceability order, so
$A_{\mathrm{LOW}}\subseteq A_{\mathrm{MEDIUM}}\subseteq
A_{\mathrm{HIGH}}$: higher compression adds LATENT units without restoring
earlier ones to text.

\noindent\textbf{Codebook construction and mixed CoT realization.}
LATENT units use a shared finite codebook \cite{oord2017vqvae}, not
span-specific placeholders. SAUs from a task-independent corpus are encoded
by frozen SpanBERT \cite{joshi2020spanbert}, followed by IncrementalPCA,
normalization, and spherical mini-batch $k$-means to form
$E=\{e_k\}_{k=1}^{K_{\mathrm{code}}}$.

A LATENT unit $S_i$, represented by the transformed, normalized vector $h_i$,
takes its nearest centroid's code:
\begin{equation}
k_i^*=\arg\max_{1\leq k\leq K_{\mathrm{code}}} h_i^\top e_k,
\qquad L_i=\texttt{<LAT\_}k_i^*\texttt{>}.
\end{equation}
\begin{samepage}
Code assignment selects the symbol, not the action. The mixed CoT is
\begin{equation}
\tilde R^{(c)}=\bigoplus_{i=1}^{N}\tilde u_i^{(c)},\qquad
\tilde u_i^{(c)}=
\begin{cases}
L_i, & i\in A_c,\\
S_i, & i\notin A_c.
\end{cases}
\end{equation}
\end{samepage}
The original order is preserved. The Student receives special tokens
\texttt{<LAT\_xxxx>} with trainable embeddings, not the offline span vectors.

\noindent\textbf{Plan-then-generate transfer.}
In Stage I, the Student predicts the serialized SAU-and-action plan
$Z_{\mathrm{plan}}^{(c)}=\operatorname{Serialize}(S_{1:N};a_{1:N}^{(c)})$
from $(q,R,c)$:
\begin{equation}
\mathcal L_{\mathrm{plan}}=-\log P_\theta\left(
Z_{\mathrm{plan}}^{(c)}\mid q,R,c\right).
\end{equation}
Stage II initializes from Stage I and removes $R$, training generation from
$(q,c)$ with $\tau_{\mathrm{FA}}=\texttt{<FINAL\_ANSWER>}$:
\begin{equation}
\mathcal L_{\mathrm{gen}}=-\log P_\theta\left(
\tilde R^{(c)},\tau_{\mathrm{FA}},y^*\mid q,c\right).
\end{equation}
Training across NONE, LOW, MEDIUM, and HIGH yields one Student that, at
deployment, generates $\tilde R^{(c)}$, $\tau_{\mathrm{FA}}$, and $\hat y$
from $(q,c)$ without the reference trace, reference answer, or Teacher.

\section{EXPERIMENTS}
\label{sec:experiments}

\noindent\textbf{Models, data, and benchmarks.}
We evaluate Qwen3-8B/14B Students \cite{yang2025qwen3} on MATH-500
\cite{hendrycks2021math,lightman2024verify}, StrategyQA \cite{geva2021strategyqa},
OlympiadBench \cite{he2024olympiadbench}, WorldTree V2 \cite{xie2020worldtree}, and
ProofWriter \cite{tafjord2021proofwriter}. Pre-answer lengths of correct,
uncompressed development CoTs define Standard-CoT (MATH-500 and StrategyQA)
and Long-CoT (the other tasks). A frozen Qwen3-235B checkpoint supplies
correct-answer traces; a frozen Qwen3-235B Teacher constructs offline
KEEP/LATENT targets. Supervision uses official training splits when available
and otherwise disjoint external/synthetic corpora; evaluation instances and
their reference solutions are excluded from target construction, codebook
fitting, and model selection. Table~\ref{tab:main-results} reports answer accuracy and achieved CR
for both Student scales.

\noindent\textbf{Baselines and evaluation protocol.}
References are No CoT (100\% CR), Raw Full CoT (uncompressed), and Ours-NONE
(same-Student, 0\% CR). Compression baselines are TokenSkip, Token Assorted,
and SPOT \cite{xia2025,su2025token,chu2026spot}, which
respectively prune text, mix text with latent tokens, and replace spans with
latent states. They share Student scales, trace pool, splits, validation-only
selection, and decoding. Ablations are dedicated Qwen3-8B studies
macro-averaged over all five benchmarks. Comparisons hold data, computation,
and evaluation fixed. Final-answer accuracy is primary. The effective reasoning length
$N_{\mathrm{eff}}$ counts each generated text token and registered latent symbol
before the final answer. For setting $m$ on benchmark $b$,
\begin{equation}
\mathrm{CR}_{m,b}=\left(1-\frac{N_{\mathrm{eff}}(m,b)}
{N_{\mathrm{eff}}(\text{Ours-NONE},b)}\right)\times100\%.
\end{equation}
CR uses the corresponding scale-and-task Ours-NONE output as denominator; Raw
Full CoT reports accuracy only. LOW, MEDIUM, and HIGH target 25\%/50\%/75\% SAU
replacement, not output CR; overall achieved CRs are about 23\%/44\%/65\%,
with task-level variation (Table~\ref{tab:main-results}). A baseline point matches when
its task-level CR is within four percentage points of SynLat;
$\dagger$ marks the nearest available point otherwise. Group and overall
results macro-average the selected task points. Main results average
independent training and generation seeds.

\begin{table}[!t]
  \centering
  \caption{Compression-unit comparison at comparable achieved CRs. Only Token
  Deletion also changes representation. Acc. Ret. (\%) is relative to each
  row's NONE accuracy; Avg. Len. counts tokens per unit. Const./Formula Split
  (\%) measures constituent/formula spans split by unit boundaries;
  CR (\%) is achieved compression rate.}
  \label{tab:allocation-interface}
  \setlength{\tabcolsep}{2.0pt}
  \renewcommand{\arraystretch}{1.05}
  \begin{tabular*}{\columnwidth}{@{\extracolsep{\fill}}lccccc@{}}
    \toprule
    & \multicolumn{3}{c}{\textbf{Unit characteristics}}
    & \multicolumn{2}{c}{\shortstack{\textbf{End-to-end}\\\textbf{result}}} \\
    \cmidrule(lr){2-4}\cmidrule(lr){5-6}
    \shortstack[l]{\textbf{Compression}\\\textbf{scheme}}
    & \shortstack{\textbf{Avg.}\\\textbf{len.}}
    & \shortstack{\textbf{Const.}\\\textbf{split $\downarrow$}}
    & \shortstack{\textbf{Formula}\\\textbf{split $\downarrow$}}
    & \shortstack{\textbf{Acc.}\\\textbf{ret. $\uparrow$}}
    & \textbf{CR} \\
    \midrule
    Token Deletion & 1.0 & 76.8 & 69.5 & 89.1 & 44.1 \\
    5-Token & 4.7 & 35.4 & 41.6 & 91.4 & 44.6 \\
    5-Token \& Guard & 4.6 & 28.9 & 15.7 & 92.2 & 44.5 \\
    Step & 18.6 & 5.7 & 7.1 & 89.6 & 43.9 \\
    \textbf{SAU (Ours)} & 3.6 & 8.1 & \textbf{5.9} & \textbf{94.0} & 44.3 \\
    \bottomrule
  \end{tabular*}
\end{table}

\section{RESULTS}
\label{sec:results}

\noindent\textbf{Main results.}
Under the reported achieved-CR selection protocol, Table~\ref{tab:main-results}
shows aggregate accuracy gains for SynLat at MEDIUM and HIGH across both
Student scales, increasing with compression.
At MEDIUM, it exceeds the strongest evaluated baseline overall by 3.6 and 2.6
percentage points for Qwen3-8B and Qwen3-14B, respectively. At HIGH, the
margins widen to 7.0 and 5.5 points. For Qwen3-8B at HIGH, the margin is 4.7
points on Standard-CoT and 8.5 points on Long-CoT. Baselines outperform
SynLat on MATH-500 for Qwen3-8B at LOW, ProofWriter
for Qwen3-14B at LOW, and StrategyQA for Qwen3-14B at MEDIUM.

\noindent\textbf{Compression-unit ablation.}
At approximately 44\% achieved CR, Table~\ref{tab:allocation-interface}
shows that SAUs retain 94.0\% of NONE accuracy. Fixed 5-token windows trail by
2.6 points, showing that a similar length cap is insufficient; guarded windows
still trail by 1.8 points, so simple boundary protection does not recover
syntax-aligned allocation. Steps trail by 4.4 points despite low split rates,
indicating that coarse units hinder selective compression. SAUs therefore best
balance formula preservation and granularity among the tested unit definitions
for mixed text--latent compression. Token Deletion is not a pure boundary control because
it also changes the compressed representation.

\noindent\textbf{Planning and training ablations.}
On fixed SAUs, answer-\linebreak conditioned planning retains 94.0\%/\allowbreak89.3\% of NONE
accuracy at MEDIUM/\allowbreak HIGH, versus 92.0\%/\allowbreak86.2\% for the strongest answer-free
alternatives. Removing action-plan supervision reduces mean accuracy across the
three compressed levels from 73.1 to 70.6, supporting the planning-and-transfer
design.

\section{CONCLUSION}
\label{sec:conclusion}

In this paper, we revisit CoT compression as a problem of allocating limited
representation fidelity. We propose SynLat, a text--latent compression
framework that bases compression decisions on Syntax-Aligned Units, allowing
local reasoning content to remain in text or be converted into latent
representations according to its role. Experiments show that, under matched
output budgets, SynLat achieves higher final-answer accuracy than
representative compression methods, with particularly large gains on tasks that
require longer reasoning traces. These results indicate that CoT compression is
not only about reducing visible tokens, but also about maintaining a better
match between reasoning structure and answer-relevant information during
compression.

\section{COMPLIANCE WITH ETHICAL STANDARDS}

This study uses publicly available benchmark datasets and does not involve
human or animal subjects. No ethical approval was required.

\section{ACKNOWLEDGMENTS}

This work was supported by the School of Electrical and Electronic Engineering,
Nanyang Technological University, through the provision of GPU computing
resources. The authors have no relevant financial or non-financial interests
to disclose.

\nocite{geva2021strategyqa,he2024olympiadbench,
hendrycks2021math,joshi2020spanbert,lightman2024verify,oord2017vqvae,
tafjord2021proofwriter,xie2020worldtree,yang2025qwen3}
\bibliographystyle{IEEEbib}
\bibliography{refs}

\end{document}